\documentclass[11pt]{article}
\usepackage{coling2018}
\usepackage{times}
\usepackage{url}
\usepackage{latexsym}

\usepackage{url}

\usepackage{float}
\usepackage{url}
\usepackage{lipsum,caption,graphicx}
\usepackage{csquotes}

\usepackage{framed}
\usepackage{fixltx2e}
\usepackage{xcolor}
\usepackage{amsmath,array,graphicx}
\usepackage{tikz}

\usepackage{enumitem}
\usepackage{multirow}

\title{Robust Lexical Features for Improved Neural Network\\Named-Entity Recognition}

\author{Abbas Ghaddar \\
	RALI-DIRO \\
	Universit\'e de Montr\'eal \\
	Montr\'eal, Canada \\
	{\tt abbas.ghaddar@umontreal.ca} \\\And
	Philippe Langlais \\
	RALI-DIRO \\
	Universit\'e de Montr\'eal \\
	Montr\'eal, Canada \\
	{\tt felipe@iro.umontreal.ca} \\}

\date{}

\newcommand{\winer}{WiNER}

\newcommand{\mightmention}[1]{}
\newcommand{\problem}[1]{\textcolor{red}{$\star$}}
\newcommand{\answer}[1]{\textcolor{blue}{$\#$}}
\newcommand{\todoreview}[1]{\textcolor{green}{$@$}}
\newcommand{\comment}[1]{}

\newcommand{\conll}{\textsc{CoNLL}}

\newcommand{\onto}{\textsc{OntoNotes}}

\newcommand{\per}{\textsc{per}}
\newcommand{\loc}{\textsc{loc}}
\newcommand{\org}{\textsc{org}}
\newcommand{\misc}{\textsc{misc}}
\newcommand{\gpe}{\textsc{gpe}}
\newcommand{\norp}{\textsc{norp}}
\newcommand{\fac}{\textsc{fac}}

\newcommand{\sskip}{\textsc{Sskip}}

\newcommand{\lr}{\textsc{LS}}

\newcommand{\wifine}{WiFiNE}

\newcommand{\felipe}[2]{}

\newcommand{\lll}{\textsc{lex}}
\newcommand{\ggg}{\textsc{gaz}}
\newcommand{\ccc}{\textsc{cap}}
\newcommand{\chhh}{\textsc{che}}
\newcommand{\eee}{\textsc{emb}}
\newcommand{\mmm}{\textsc{lme}}
\newcommand{\sss}{\textsc{ls}}

\begin{document}
	\maketitle
	\begin{abstract}
		Neural network approaches to Named-Entity Recognition reduce the need for carefully hand-crafted features. While some features do remain in state-of-the-art systems, lexical features have been mostly discarded, with the exception of gazetteers. In this work, we show that this is unfair: lexical features are actually quite useful. We  propose to embed words and entity types into a low-dimensional vector space we train from annotated data produced by distant supervision thanks to Wikipedia. From this, we compute --- offline --- a feature vector representing each word. When used with a vanilla recurrent neural network model, this representation yields substantial improvements. We establish a new state-of-the-art  F1 score of 87.95 on  \onto{ 5.0}, while matching state-of-the-art performance with a F1 score of 91.73 on the over-studied \conll{-2003} dataset.

	\end{abstract}
	
	\section{Introduction}
	%
	%
	
	\blfootnote{
		%
		%
		%
		%
		%
		%
		\hspace{-0.65cm}  
		This work is licensed under a Creative Commons 
		Attribution 4.0 International License.
		License details:
		\url{http://creativecommons.org/licenses/by/4.0/}
	}

	Named-Entity Recognition (NER) is the task of identifying textual mentions and classifying them into a predefined set of types.   Various approaches have been proposed to tackle the task, from hand-crafted feature-based machine learning models like conditional random fields~\cite{finkel2005incorporating} and perceptron~\cite{ratinov2009design}, to deep neural models~\cite{collobert2011natural,ma2016end,strubell2017fast}.
	
	Word representations~\cite{turian2010word,mikolov2013distributed}, also known as word embeddings, are a key element for multiple NLP tasks including NER~\cite{collobert2011natural}. Due to the small amount of named-entity annotated data, embeddings are used to extend, rather than replace, hand-crafted features in order to obtain state-of-the-art performance~\cite{lample2016neural}. 
	Recent studies~\cite{yang2017neural,sogaard2016deep} have explored methods for supplying deep sequential taggers with complementary features to standard embeddings. \newcite{peters2017semi} and \newcite{tran2017named} tested special embeddings extracted from a neural language model (LM) trained on a large corpus. LM embeddings capture context-dependent aspects of word meaning using future (forward LM) and previous (backward LM) context words. When this information is added to standard features, it leads to significant improvements in NER. 	Also, \newcite{chiu2015named} showed that external knowledge resources (namely gazetteers) are crucial to NER performance. Gazetteer features encode the presence of word $n$-grams in predefined lists of NEs. 
	
	In this work, we discuss some of the limitations of gazetteer features and propose an alternative lexical representation which is trained offline and that can  be added to any neural NER system. In a nutshell, we embed words and entity types into a joint vector space by leveraging  \wifine~\cite{ghaddara2018wifine}, a ressource which automatically annotates mentions in Wikipedia with 120 entity types. From  this vector space, we compute for each word a  120-dimensional vector, where each dimension encodes the similarity of the word with an entity type. We call this vector an \lr{} representation, for Lexical Similarity. When included in a vanilla LSTM-CRF NER model, \lr{} representations lead to significant gains.  We establish a new state-of-the-art F1 score of  87.95 on \onto{ 5.0}, while matching state-of-the-art performance on the over-studied \conll{-2003} dataset. \felipe{, without making use of LM embeddings as features.}{Unsure this is claim you want to make: because it raises the question: why the hell not?}

	In the rest of this paper, we motivate our work in Section~\ref{sec:motivation}. We describe how we compute \lr{} vectors in Section~\ref{sec:method}. We present our system in Section~\ref{sec:system} and report results in Section~\ref{sec:Experiments}. In Section~\ref{sec:related}, we discuss related works before concluding in Section~\ref{sec:conclusion}.
	
	\section{Motivation}
	\label{sec:motivation}
	
	Gazetteers are lists of entities that are associated with specific NE categories. They are widely used as a feature source in NER, and have been successfully included in  feature-based~\cite{ratinov2009design} and neural~\cite{chiu2015named} models. Typically, lists of entities are compiled from structured data sources such as DBpedia~\cite{auer2007dbpedia} or Freebase~\cite{Bollacker:2008}. The surface form of the title of a Wikipedia article, as well as aliases and redirects are mapped to an entity type using the \texttt{object\_type} attribute of the related DBpedia (or Freebase) page. \newcite{ratinov2009design} use this methodology to compile 30 lists of fine-grained entity types extracted from Wikipedia, while \newcite{chiu2015named} create 4 gazetteers that map to CoNLL categories (\per, \loc, \org{} and \misc). Despite their importance, gazetteer-based features suffer from a number of limitations.

	\begin{itemize}
		
		\item \textbf{Binary representation.} Gazetteer features encode only the presence of an $n$-gram in each list and omit its relative frequency. For example, the word \enquote{France} can be used as a person, an organization, or a location, while it likely refers to the country most of the time. Binary features cannot capture this preference. 
		
		\item \textbf{Generation.} At test time, we need to match every n-gram (up to the length of the longest lexicon entry) in a sentence against entries in the lexicons, which is time consuming. In their work, \newcite{chiu2015named} use 4 lists that count over 2.3M entries.
		
		\item \textbf{Non-entity words.}	Gazetteer features do not capture signal from non-entity words, while earlier feature-based models strived to encode  that some words (or $n$-grams) trigger specific entity types.  For instance, words such as \enquote{eat}, \enquote{directed} or \enquote{born} are words that typically appear after a mention of type \per.
		
	\end{itemize}

	To overcome those limitations, we propose an alternative approach where we embed annotations mined from Wikipedia into a vector space from which we compute a feature vector that represent words. This vector compactly and efficiently encodes both gazetteer and lexical information. Note that at test time, we only have to feed our model with this feature vector, which is efficient.

	\section{Our Method}		
	\label{sec:method}
	

	\subsection{Embedding Words and Entity Types}
	
	Turning Wikipedia into a corpus of named-entities annotated with types is a task that received continuous attention over the years~\cite{nothman2008transforming,al2015polyglot,ghaddar2017winer}. It consists mainly in exploiting the hyperlink structure of Wikipedia in order to detect entity mentions. Then, structured data from a knowledge base (for instance Freebase) are used to map hyperlinks to entity types. Because the number of anchored strings in Wikipedia is no more than 3\% of the text tokens\felipe{REFERENCE?}{il me semble qu'il y a avait une reference qui disait un truc comme cela}, \newcite{ghaddar2017winer} proposed to augment Wikipedia articles with  mentions unmarked in Wikipedia, thanks to a mix of heuristics that benefit the Wikipedia structure~\cite{ghaddar2016coreference}, as well as a coreference resolution system adapted specifically to Wikipedia~\cite{GHADDAR16.192}.
	
	The authors applied their approach on English Wikipedia and produce coarse (4 classes) and fine-grained (120 labels) named-entity annotations, leading to \winer~\cite{ghaddar2017winer} and \wifine~\cite{ghaddara2018wifine}. In this work, we adopt \wifine{ } which is publicly available at \url{http://rali.iro.umontreal.ca/rali/en/wifiner-wikipedia-for-et} as our source of annotations. Each entity mention is mapped (via its Freebase \texttt{object\_type} attribute) to a pre-defined set of 120 entity types. Types are stored in a 2-level hierarchical structure (e.g. \texttt{/person} and \texttt{/person/musician}). The corpus consist of 3.2M Wikipedia articles, comprising 1.3G  tokens that we annotated with 157.4M named-entity mentions and their types.
	We used this very large quantity of automatically annotated data for jointly embedding words and entity types into the same low-dimensional space.  The key idea consists in learning an embedding for each entity type using its surrounding words. For instance, the embedding for \texttt{/product/software} will be trained using context words that surround all entities that were (automatically) labelled as \texttt{/product/software} in Wikipedia. In practice, we found that simply concatenating a sentence (v1) with its annotated version (v2),  as illustrated in Figure~\ref{fig:corpus}, offers a simple but efficient way of combining words and entity types so that embeddings can make good use of them.
	
	\begin{figure}[!h]
		
		\begin{framed}
			\textbf{(v1)} On \textbf{October 9, 2009}, the \textbf{Norwegian Nobel Committee} announced that \textbf{Obama} had won the \textbf{2009 Nobel Peace Prize}.
			
			\textbf{(v2)} On \texttt{/date}, the \texttt{/organization/government\_agency} announced that \texttt{/person/politician} had won the \texttt{/award}.
		\end{framed}
		
		\caption{Example of the two variants of a given sentence.}
		\label{fig:corpus}
	\end{figure}


	We use the FastText toolkit~\cite{bojanowski2016enriching} to learn the uncased embeddings for both words and entity types. We train a skipgram model to learn 100-dimensional vectors with a minimum word frequency cutoff of 5, and a window size of 5. This configuration (recommended by the authors) performs the best in the experiments described in Section~\ref{sec:Experiments}. Since FastText learns representations of character \textit{n}-grams, it has the ability to produce vectors for unknown words.

	\begin{figure*}[h]
		\begin{center} 
			\includegraphics[scale=0.33]{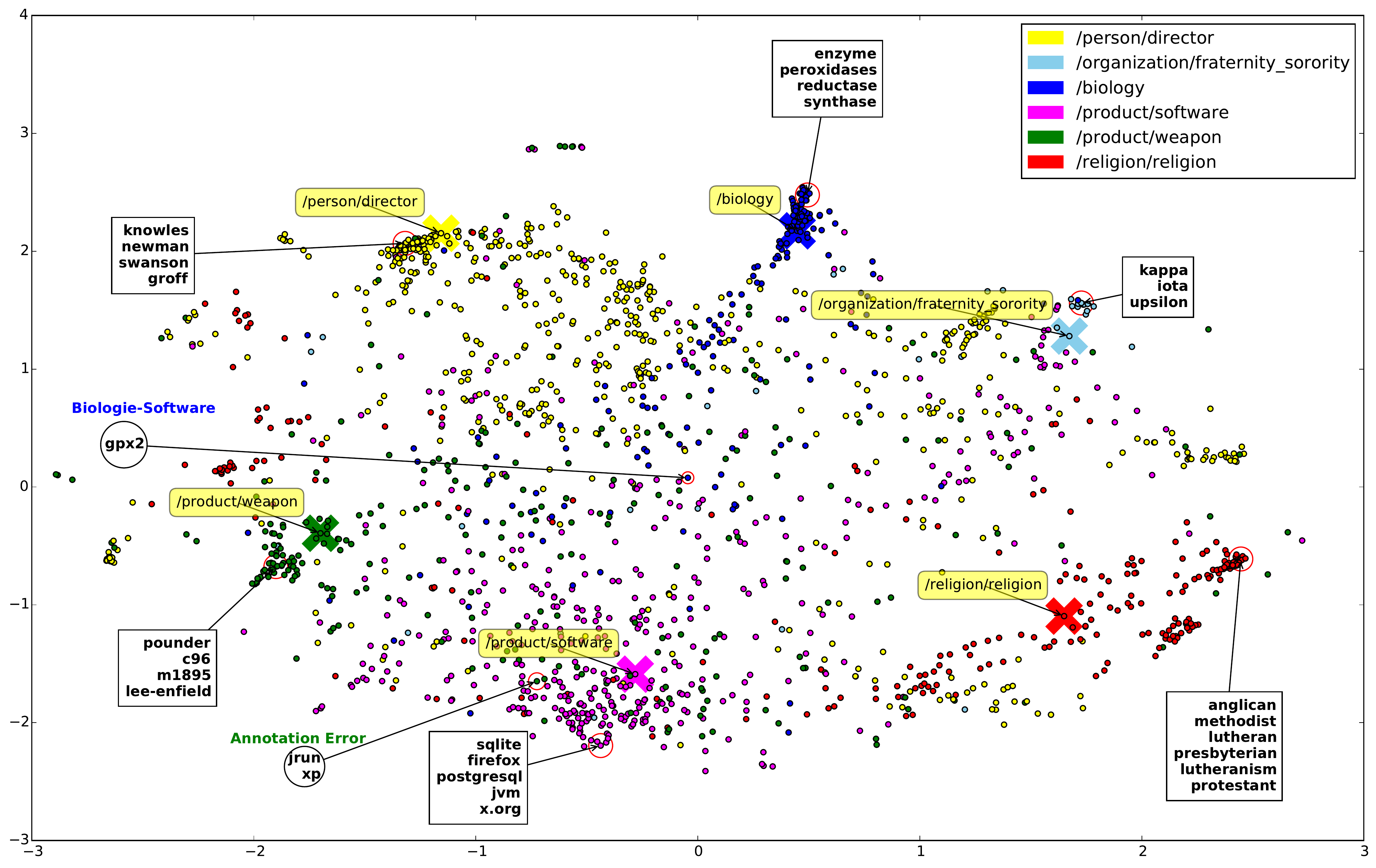} 
			
			\caption{Two-dimensional representation of the vector space which embeds both words and entity types. Big Xs indicate entity types, while circles refer to words (i.e. named-entities, here).}
			
			\label{fig.tnse}
		\end{center}
	\end{figure*}
			
	Figure~\ref{fig.tnse} illustrates a T-SNE~\cite{tsne} two-dimensional projection of the embedding of 6 entity types and a sample of 1500 words. Entity type embeddings are marked by big Xs, while circles indicate words. For visualization proposes, we only plot single-word mentions that were annotated in \wifine{ } with one of those 6 types. Words were randomly and proportionally sampled according to the frequency of each entity type. In addition, words have the color associated with the most frequent type they were annotated with in \wifine.

	We observe that mentions often annotated by a given type in our resource tend to cluster around this entity type. For instance, \enquote{firefox} is close to the type \texttt{/product/software}, while \enquote{enzyme} is close to the \texttt{/biology} entity type. We also notice that words that are labelled with different types tend to appear between types they were annotated with. For instance,  \enquote{gpx2}, which is used both as a software and as a gene, has its embedding in between \texttt{/product/software} and \texttt{/biology}. 	
	We inspected some of the words plotted in Figure~\ref{fig.tnse}, and found that \enquote{jrun} and \enquote{xp} are incorrectly labelled as \texttt{/product/weapon} in \wifine. But since these words are seen in a \textit{software} context,  their embeddings are closer to the \texttt{/product/software} embedding than the \texttt{/product/weapon} one. We feel this tolerance to noise is a desirable feature, one that hopefully allows a more efficient use of distant supervision. 
	Last, we also observe the tendency of rare words to cluster around their entity type. For instance,  \enquote{iota} and \enquote{x.org} are embedded near their respective types, despite the fact that they appear less than 30 times in the version of Wikipedia used to compile \wifine.
	
	\subsection{\lr{} Representation}
	
	This joint vector space only serves the purpose of associating to each word a \lr{} representation, that is, a 120-dimensional vector where the $i$th coefficient is a value in the $[-1,+1]$ interval, equal to the cosine similarity\footnote{The cosine similarity outperforms other metrics in our experiments.} between the word embedding  and the embedding of the $i$th entity type (we have 120 types).

	\begin{table}[!h]
		\begin{center}
			\begin{tabular}{|l|l|l||l|l|l|}
				\hline 
				\bf Word   & \bf Entity Type & \bf Sim & \bf Word   & \bf Entity Type & \bf Sim  \\ 
				\hline
				\multirow{3}{*}{hilton} & \texttt{/building/hotel} & 0.58 & \multirow{3}{*}{located} & \texttt{/location} &  0.47\\
				& \texttt{/building/restaurant} & 0.46 & & \texttt{/location/city} & 0.44\\
				& \texttt{/person/actor} & 0.37 & & \texttt{/building} & 0.40  \\
				\hline	
				\multirow{2}{*}{gpx2} & \texttt{/biology} & 0.69 & \multirow{2}{*}{directed} & \texttt{/person/director} & 0.60 \\
				& \texttt{/product/software} & 0.56 & & \texttt{/art/film} & 0.55\\
				\hline					
				\multirow{2}{*}{jrun} & \texttt{/product/software} & 0.64  & \multirow{2}{*}{in} & \texttt{/date} & 0.58\\
				& \texttt{/product/weapon} & 0.23 & & \texttt{/location/city} & 0.54\\
				\hline
				\multirow{2}{*}{dammstadt} & \texttt{/location/city} & 0.45 & \multirow{2}{*}{won} & \texttt{/award} & 0.53 \\
				& \texttt{/location/railway} & 0.44 & & \texttt{/event/sports\_event} & 0.53 \\	
				\hline
				
			\end{tabular}
			\caption{Topmost similar entity types to a few single-word mentions (left table) and non-entity words (right table).}
			\label{tab.sim}
		\end{center}
	\end{table}

	Table~\ref{tab.sim} shows the topmost similar entity types for proper names (left column) and common words (right column). We observe that ambiguous mentions (those annotated with several types) are adequately handled. For instance, the \lr{} representation of the word \enquote{hilton} encodes that it more often refers to a hotel or a restaurant than to an actress. Also, we observe that entity words that are either not or rarely annotated in \wifine{} are still adequately associated with their right type. For instance, \enquote{dammstadt}, which appears only 5 times in \wifine, and which refers to the Damm city in Germany, is most similar to \texttt{/location/city} and \texttt{/location/railway}. Interestingly, this mention does not have its page in English Wikipedia.	
	Furthermore, we observe that non-entity context words have a strong similarity to types they precede or succeed. For instance the verb \enquote{directed} is very close to \texttt{/person/director}, an entity type that usually precedes it, and to \texttt{/art/film}, that usually follows it. Likewise, the preposition \enquote{in} is near \texttt{/date} and \texttt{/location/city}, which frequently follow "in".

	\subsection{Strength of the \lr{} Representation}

	To summarize, we propose a compact lexical representation which is computed offline, therefore incurring no computation burden at test time This representation encodes the preference of an entity-mention word for a given type, an information out of reach of binary gazetteer features. It also lends itself nicely to the inclusion of lexical features that have been successfully used in earlier feature-based systems~\cite{ratinov2009design,luo-2015}. Also, because entity types are well represented in \wifine, their embeddings are robust: Our representation does accommodate unfrequent words and seems tolerant to the inherent noise of distant supervision.
	
	\section{Our NER System}
	\label{sec:system}
	
	In order to test the efficiency of our lexical feature representation, we implemented a state-of-the-art NER system we now describe.
	
	\subsection{Bi-LSTM-CRF Model}
	\label{sec:Bi-LSTM-CRF Model}

\begin{figure*}[hbt]
		\hspace*{-4mm}
		
		\begin{minipage}[c]{0.5\linewidth}
			\hspace{1cm} \includegraphics[scale=0.35]{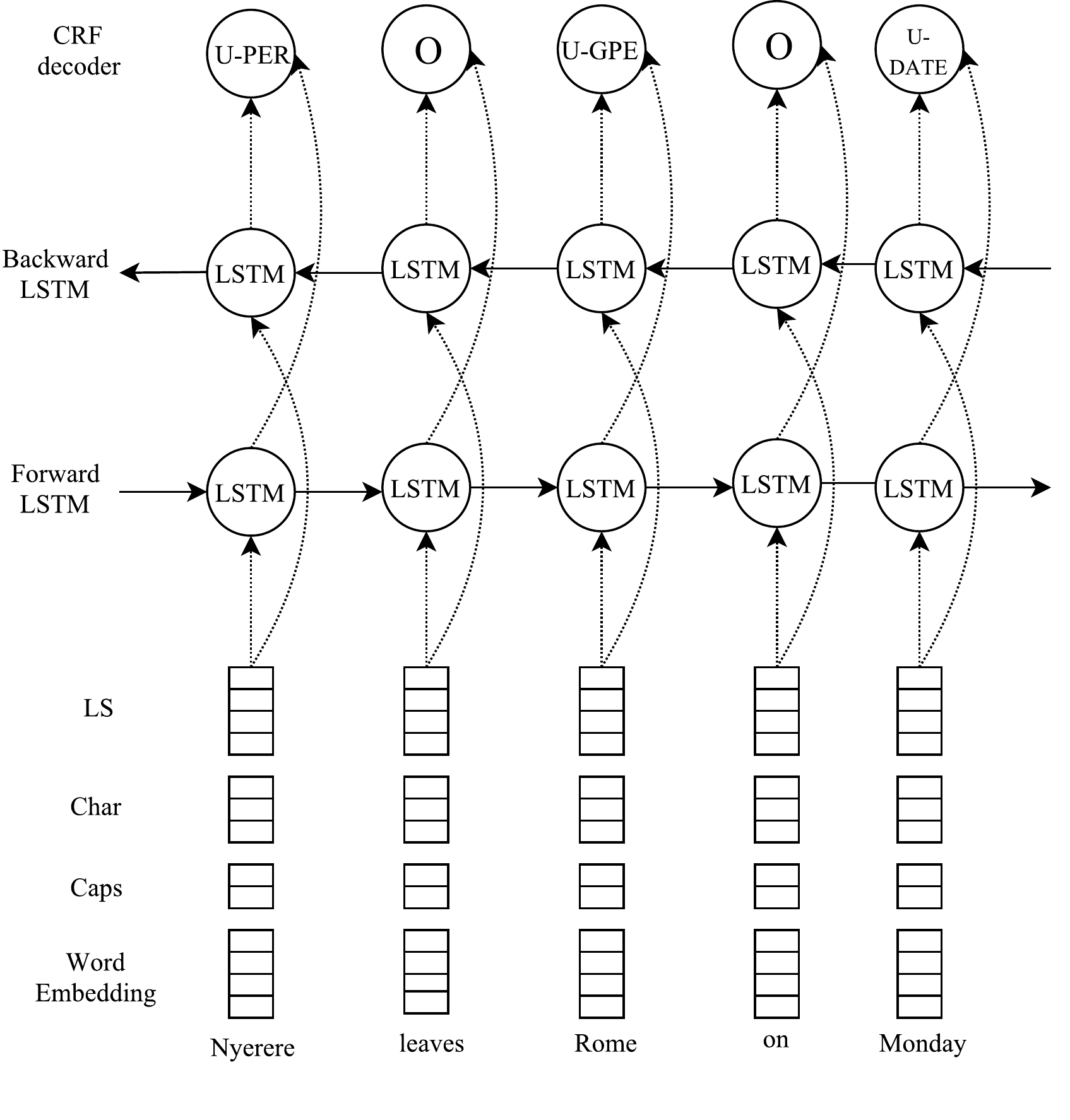}
		\end{minipage}
		\begin{minipage}[c]{0.5\linewidth}
			\includegraphics[scale=0.45]{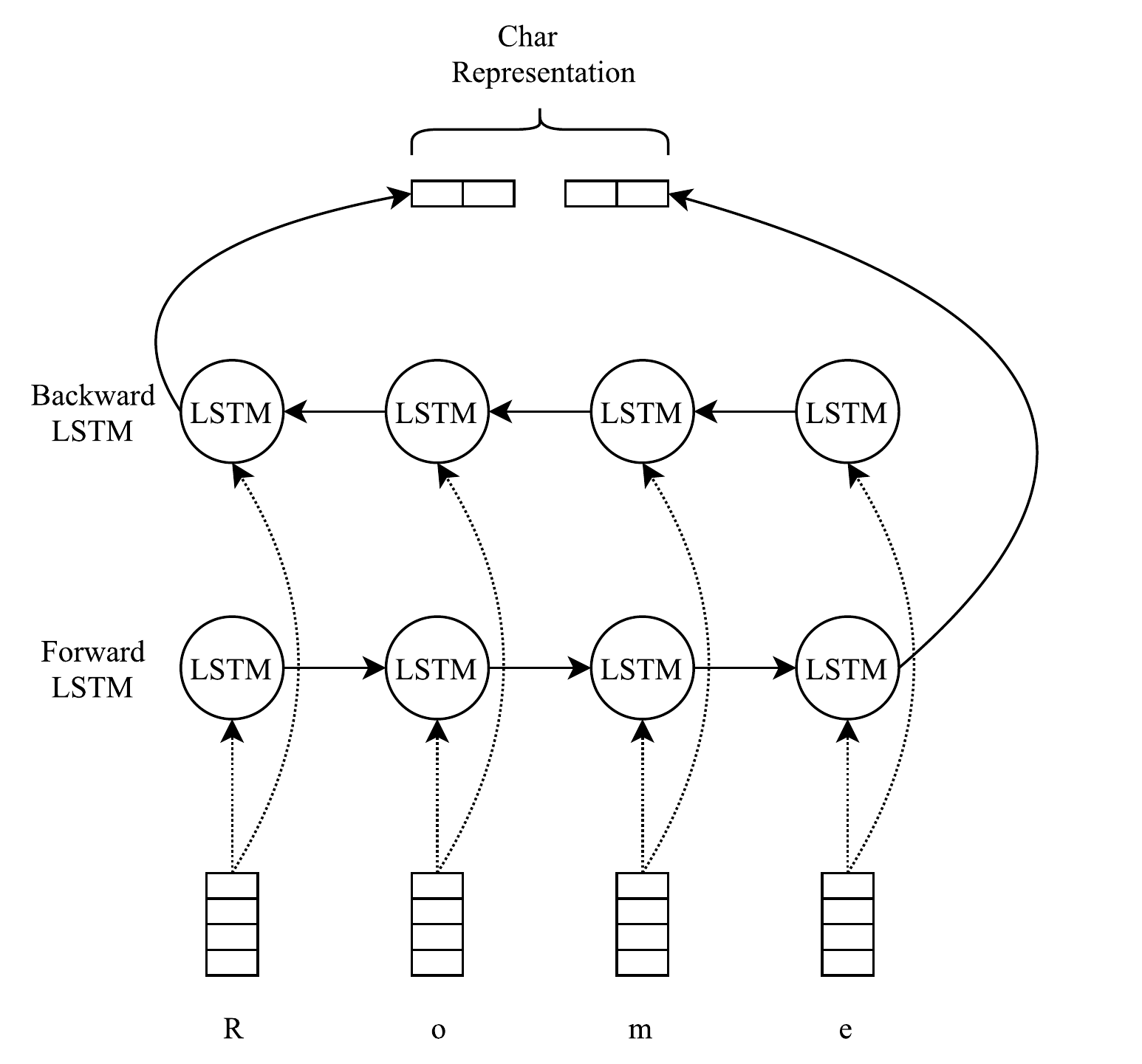}
		\end{minipage}
		
		\caption{\textbf{Left Figure:} Main architecture of our NER system. \textbf{Right Figure:} Character representation of the word \enquote{Roma} given to the word-level bi-LSTM.}
		
		\label{fig:words.lstm}
		
	\end{figure*}

	We adopt the popular Bi-LSTM-CRF architecture (Figure~\ref{fig:words.lstm}), a \textit{de facto} baseline in many sequential tagging tasks~\cite{lample2016neural,sogaard2016deep,chiu2015named}.

	\subsection{Features}
	
	In addition to the \lr{} vector, we incorporate publicly available pre-trained embeddings, as well as character-level, and capitalization features. Those features have been shown to be crucial for state-of-the-art performance. 
	
	\subsubsection{Word Embeddings}
	\label{sec:pretrained}
	We experimented with several publicly available word embeddings, such as Senna~\cite{collobert2011natural}, Word2Vec~\cite{mikolov2013distributed}, GloVe~\cite{pennington2014glove}, and \sskip~\cite{yulia2015not}. We find that the latter  performs the best in our experiments. \sskip{} embeddings  are 100-dimensional case sensitive vectors that where trained using a \textit{n}-skip-gram model~\cite{yulia2015not} on 42B tokens. These embeddings were previously used by ~\cite{lample2016neural,strubell2017fast}, who report good performance on \conll{}, and state-of-the-art results on \onto{} respectively. Note that these pre-trained embeddings are adjusted during training.

	\subsubsection{Character Embeddings}
	\label{sec:Chars}

	Following  \cite{lample2016neural}, we use a forward and a backward LSTM to derive a representation of each word from its characters (right part of Figure~\ref{fig:words.lstm}). A character lookup table is randomly initialized, then trained at the same time as the Bi-LSTM model sketched  in Section~\ref{sec:Bi-LSTM-CRF Model}.
	
	\subsubsection{Capitalization Features}
	\label{sec:Caps}
	
	Similarly to previous works, we use capitalization features for characterizing certain categories of capitalization patterns: \texttt{allUpper}, \texttt{allLower}, \texttt{upperFirst}, \texttt{upperNotFirst} , \texttt{numeric} or \texttt{noAlphaNum}. We define a random lookup table for these features, and learn its parameters during training.
	
	\subsubsection{\lr{} Vectors}
	
	Contrarily to previous features, lexical vectors are computed offline and are not adjusted during training. 
	We found useful in practice to apply a \texttt{MinMax} scaler in the range $[-1,+1]$ to each \lr{} vector we computed; thus, $[.., 0.095, .., 0.20,.., 0.76,..]$  becomes $[..,-1,..,-0.67,..,1,..]$.
	
	\section{Experiments}
	\label{sec:Experiments}

	\subsection{Data and Evaluation}
	\label{sec:Datasets}
	
	We consider two well-established NER benchmarks: \conll{-2003} and \onto{ 5.0}. Table~\ref{tab.dataset} provides an overview of the two datasets. As we can see, \onto{} is much larger. 
	For both datasets, we convert the \texttt{IOB} encoding to \texttt{BILOU}, since~\newcite{ratinov2009design} found the latter to perform better. In keeping with others, we report mention-level F1 score using the \texttt{conlleval} script\footnote{\url{http://www.cnts.ua.ac.be/conll2000/chunking/conlleval.txt}}.

	The \conll{-2003} NER  dataset~\cite{tjong2003introduction} is a well known collection of Reuters newswire articles that contains a large portion of sports news. It is annotated with four entity types: \textit{Person (\per), Location (\loc), Organization (\org) and Miscellaneous (\misc)}. The four entity types are fairly evenly distributed, and the train/dev/test datasets present a similar type distribution. 
	
	\begin{table}[h]
		\begin{center} 
			
			\setlength\tabcolsep{5pt} 
			
			\begin{tabular}{clrrrl}
				
				\textbf{Dataset} && \textbf{Train} & \textbf{Dev} & \textbf{Test}\\
				\hline
				\conll{-2003} & \textit{\#tok} & 204,567 & 51,578 & 46,666 \\
				& \textit{\#ent} & 23,499 & 5,942 & 5,648\\ 
				\hline
				\onto{ 5.0} & \textit{\#tok} & 1,088,503 & 147,724 & 152,728 \\
				& \textit{\#ent} &  81,828 & 11,066 & 11,257 \\	 
				\hline
			\end{tabular} 
		\end{center}
		
		\caption{Statistics of the \conll{-2003} and \onto{ 5.0} datasets. \textit{\#tok} stands for the number of tokens, and \textit{\#ent} indicates the number of named-entities gold annotated. }
		\label{tab.dataset} 
		
	\end{table}
	
	The \onto{ 5.0}  dataset~\cite{hovy2006ontonotes,pradhan2013towards}  includes texts from five different genres: broadcast conversation (200k), broadcast news (200k), magazine (120k), newswire (625k), and web data (300k). This dataset is annotated with 18 entity types, and is much larger than \conll. Following previous researches~\cite{chiu2015named,strubell2017fast}, we use the official train/dev/test split of the CoNLL-2012 shared task~\cite{pradhan_conll-2012_2012}.  Also, we exclude (both during training and testing) the New Testaments portion as it does not contain gold NE annotations.

	\subsection{Training and Implementation}
	
	Training is carried out by mini-batch stochastic gradient descent (SGD) with a momentum of 0.9 and a gradient clipping of 5.0. The mini-batch is 10 for both datasets, and learning rates are 0.009 and 0.013 for \conll{} and \onto{} respectively. More sophisticated optimization algorithms such as AdaDelta~\cite{zeiler2012adadelta} or Adam~\cite{kingma2014adam} converge faster, but none outperformed  SGD with exponential learning rate decay in our experiments.

	Our system uses a single Bi-LSTM layer at the word level whose hidden dimensions are set to 128 and 256 for \conll{} and \onto{} respectively. 	For both models, the character embedding size was set to 25, and the hidden dimension of the forward and backward character LSTMs are set to 50. To mitigate overfitting, we apply a dropout mask~\cite{srivastava2014dropout} with a probability of 0.5 on the input and output vectors of the Bi-LSTM layer. For both datasets, we set the dimension of capitalization embeddings to 25 and trained the models up to 50 epochs.
	
	We tuned the hyper-parameters by grid search, and used early stopping based on the performance on the development set. We varied dropout ($[0.25, 0.5, 0.65]$), hidden units ($[50, 128, 256, 300]$), capitalization ($[10, 20, 30]$) and char ($[25, 50, 100]$) embedding dimensions, learning rate ($[0.001, 0.015]$ by step $0.002$), and optimization algorithms and fixed the other hyper-parameters. We implemented our system using the Tensorflow~\cite{abadi2016tensorflow} library, and ran our models on a GeForce GTX TITAN Xp GPU. Training requires about 2.5 hours for \conll{} and 8 hours for \onto{}. 
	
	\subsection{Results on the Development Set} 
	\label{sec:dev}

	Table~\ref{tab.dev} shows the development set performance of our final models on each dataset compared to the work of~\newcite{chiu2015named}. The authors use an architecture similar to ours, but use a binary gazetteer feature set, while we use our \lr{}  representation. Since our systems involve random initialization, we report the mean as well as the standard deviation over five runs. The improvements yielded by our model on the \conll{} dataset are significant although modest, while those observed on \onto{} are more substantial. We also observe a lower variance of our system over the 5 runs.

	\begin{table}[h]
		\begin{center} 
			\begin{tabular}{|c|l|l|}
				\hline 
				& \textbf{\conll} & \textbf{\onto}\\
				\hline
				\cite{chiu2015named}  & 94.03 ($\pm$ 0.23) & 84.57 ($\pm$ 0.27)\\
				Our model & \bf 94.80 ($\pm$ 0.10) & \bf 86.44 ($\pm$ 0.14) \\
				\hline
			\end{tabular}
		\end{center}

		\caption{Development set F1 scores of our best hyper-parameter setting compared to the results reported in \cite{chiu2015named}.}
		\label{tab.dev} 
		
	\end{table}

	\subsection{Results on \conll}
	
	Table~\ref{tab:conll.res} reports our model's performance\footnote{Standard deviation on the test set is reported in Table~\ref{tab:ablation}} on the \conll{ }test set, as well as the performance of systems previously tested on this test set (the figures are those published by the authors). Because of the small size of the training set, some authors ~\cite{chiu2015named,yang2017neural,peters2017semi,peters2018deep} incorporated the development set as a part of training data after tuning the hyper-parameters. Consequently, their results are not directly comparable, so we do not report them.
	
	\begin{table}[!h]
		\begin{center}
			\setlength{\tabcolsep}{1mm}  
			\begin{tabular}{|l|lllllll|c|}
				\hline \textbf{Model} & \lll & \ggg & \ccc &\eee & \chhh & \mmm & \sss  & \textbf{F1} \\ 
				\hline
				\cite{finkel2005incorporating} 		& \textsc{+}& \textsc{+}& \textsc{+} & $\bullet$&$\bullet$&$\bullet$& $\bullet$	& 86.86 \\
				\cite{ratinov2009design} 			& \textsc{+}& \textsc{+}& \textsc{+} & $\bullet$&$\bullet$&$\bullet$&$\bullet$	& 90.88 \\
				\cite{lin2009phrase} 				& \textsc{+}& \textsc{+}& \textsc{+} & $\bullet$&$\bullet$&$\bullet$&$\bullet$	& 90.90\\
				\cite{luo-2015} &\textsc{+}& \textsc{+}& \textsc{+} & $\bullet$&$\bullet$&$\bullet$&$\bullet$	& 91.20 \\
				
				\hline
				\cite{collobert2011natural} 		& $\bullet$ & \textsc{+}& \textsc{+}& \textsc{+} &$\bullet$&$\bullet$&$\bullet$	& 89.56\\
				\cite{huang2015bidirectional} 		& $\bullet$&$\bullet$ & \textsc{+}& \textsc{+}& \textsc{+} & $\bullet$ & $\bullet$	& 90.10\\
				\cite{lample2016neural} 			& $\bullet$&$\bullet$ & \textsc{+}& \textsc{+}& \textsc{+} & $\bullet$ & $\bullet$	& 90.94 \\
				\cite{ma2016end} 				& $\bullet$&$\bullet$ & \textsc{+}& \textsc{+}& \textsc{+} & $\bullet$ & $\bullet$	& 91.21\\	 
				\cite{shen2017deep} 			& $\bullet$&$\bullet$ & \textsc{+}& \textsc{+} & $\bullet$ & $\bullet$ & $\bullet$	& 90.89 \\
				\cite{strubell2017fast} 			& $\bullet$&$\bullet$ & \textsc{+}& \textsc{+} & $\bullet$ & $\bullet$ & $\bullet$	& 90.54 \\				
				\cite{tran2017named}  			& $\bullet$&$\bullet$ & \textsc{+}& \textsc{+} & \textsc{+}& \textsc{+} & $\bullet$ 	& 91.69 \\
				\cite{liu2017empower} 			& $\bullet$&$\bullet$ & \textsc{+}& \textsc{+} & \textsc{+}& \textsc{+} & $\bullet$ 	&  91.71 \\
				\hline					
				\textbf{This work} 				& $\bullet$& \textsc{+}& \textsc{+} & \textsc{+}& \textsc{+} & $\bullet$ 	& \textsc{+}	& \bf 91.73 \\
				\hline

			\end{tabular}
		\end{center}

		\caption{F1 scores on the \conll{} test set. The first four systems are feature-based, the others are neuronal. The   feature configuration of each system is encoded with:  \lll{} which stands for \textsc{lex}ical feature, \ggg{} for \textsc{gaz}etteers, \ccc{} for \textsc{cap}italization, \eee{} for pre-trained \textsc{emb}eddings, \chhh{} for \textsc{ch}aracter \textsc{e}mbeddings, \textsc{lme} for \textsc{l}anguage \textsc{m}odel \textsc{e}mbeddings, and  \sss{} for the proposed \lr{~} feature representation. \textsc{+} indicates that the model uses the feature set.}
		
		\label{tab:conll.res} 
		
	\end{table}
		
	First, we observe that our model significantly  outperforms models that use extensive sets of hand-crafted features \cite{ratinov2009design,lin2009phrase} as well as the system of \cite{luo-2015} that uses NE and Entity Linking annotations to jointly optimize the performance on both tasks.  Second, our model outperforms as well other NN models that only use standard word embeddings, which indicates that our lexical feature vector is complementary to standard word embeddings. 
	Third, our system matches state-of-the-art performances of models that use either more complex architectures or more elaborate features. \newcite{tran2017named} use three layers of stacked residual RNN (Bi-LSTM) with bias decoding. Our model is much simpler and faster. They report a performance of 90.43 when using an architecture similar to ours. The two systems that have slightly higher F1 scores on the \conll{} dataset both use embeddings obtained from a forward and a backward Language Model trained on the One Billion Word Benchmark~\cite{chelba2013one}. They report gains between 0.8 and 1.2 points by using such LM embeddings, which suggests that \lr{} vectors are indeed efficient. Unfortunately, due to time and resource constraints,\footnote{LM embeddings are not publicly available, and according to \newcite{jozefowicz2016exploring}, they require three weeks to train on 32 GPUs.} we were not able to measure whether both features complement each other. This is left for future investigations. 
	
	\subsection{Results on \onto}	
	
	Table~\ref{tab:onto.res} reports the F1 score of our system compared to the performance reported by others on the \onto{} test set. To the best of our knowledge, we surpass previously reported F1 scores on this dataset. In particular, our system significantly outperforms the Bi-LSTM-CNN-CRF  models of \cite{chiu2015named} and \cite{strubell2017fast} by an absolute gain of 1.68 and 0.96 points respectively.  Less surprisingly, it surpasses systems with hand-crafted features, including \newcite{ratinov2009design} that use gazetteers, and the system of \newcite{durrett2014joint} which uses coreference annotation in \onto{} to jointly model NER, entity linking, and coreference resolution tasks.

	\begin{table}[h]
		
		\begin{center}
			\setlength{\tabcolsep}{1mm}  
			\begin{tabular}{|l|lllllll|c|}
				\hline \textbf{Model} & \lll & \ggg & \ccc &\eee & \chhh & \mmm & \sss & \textbf{F1} \\ 
				\hline
				\cite{finkel2009joint} 						& \textsc{+}& \textsc{+}& \textsc{+} & $\bullet$&$\bullet$&$\bullet$&$\bullet$	& 82.42 \\
				\cite{ratinov2009design}  		& \textsc{+}& \textsc{+}& \textsc{+} & $\bullet$&$\bullet$&$\bullet$&$\bullet$	& 84.88 \\
				\cite{passos2014lexicon} 				        & \textsc{+}& \textsc{+}& \textsc{+} & $\bullet$&$\bullet$&$\bullet$&$\bullet$	& 82.24 \\
				\cite{durrett2014joint} 					& \textsc{+}& \textsc{+}& \textsc{+} & $\bullet$&$\bullet$&$\bullet$&$\bullet$	& 84.04 \\
				\hline 
				\cite{chiu2015named} & $\bullet$ & \textsc{+}& \textsc{+}& \textsc{+} & \textsc{+} &$\bullet$&$\bullet$	& 86.28 \\
				\cite{shen2017deep} & $\bullet$&$\bullet$ & \textsc{+}& \textsc{+}& \textsc{+} & $\bullet$ & $\bullet$	& 86.52 \\
				\cite{strubell2017fast} & $\bullet$&$\bullet$ & \textsc{+}& \textsc{+}& \textsc{+} & $\bullet$ & $\bullet$	& 86.99 \\
				
				\hline
				\textbf{This work} 	& $\bullet$& \textsc{+}& \textsc{+} & \textsc{+}& \textsc{+} & $\bullet$ 	& \textsc{+}	& \bf 87.95 \\
				\hline
			\end{tabular}
			\footnotetext{Results as reported in \cite{RedmanSaRo17} as this data split did not exist at the time of publication.}		
		\end{center}

		\caption{F1 scores on the \onto{} test set. The first four systems are feature-based, the following ones are neuronal. See Table~\ref{tab:conll.res} for an explanation of the column of features.}
		\label{tab:onto.res} 
		
	\end{table}

	The \onto{} benchmark  is annotated with 18 types (e.g. \textsc{law}, \textsc{product}) and  contains many rare words, especially in the Web data collection. \newcite{chiu2015named} note that the 4-class gazetteer they used  yielded marginal improvements on \onto{}, contrarily to \conll. In particular, they observe that mentions that match \loc{~} entries in their gazetteer often match  \gpe, \norp{} and \fac{} lists.  
	
	\begin{table}[!h]
		\setlength{\tabcolsep}{2mm} 
		\begin{center}
			\begin{tabular}{|l|l|l|l|l|l|l|}
				\hline \bf Model & \bf BC & \bf BN & \bf MZ & \bf NW & \bf TC  & \bf WB \\
				\hline
				\cite{finkel2009joint} & 78.66 & 87.29 & 82.45 & 85.50 & 67.27 & 72.56 \\
				\cite{durrett2014joint} & 78.88 & 87.39 & 82.46 & 87.60 & 72.68 & 76.17 \\
				\cite{chiu2015named} & 85.23 & 89.93 & 84.45 & 88.39 & 72.39 & 78.38 \\
				\hline
				\textbf{This work} & \bf 86.33 & \bf 90.46 & \bf 85.91 & \bf 89.75 & \bf 75.41 & \bf 80.39 \\
				\hline
			\end{tabular}
		\end{center}
		
		\caption{Per-genre F1 scores on \onto{} (numbers taken from \newcite{chiu2015named}). BC = broadcast conversation, BN = broadcast news, MZ = magazine, NW = newswire, TC = telephone conversation, WB = blogs and newsgroups.}
		\label{tab:onto.genre} 
		
	\end{table}

	They suggest that a finer-grained gazetteer could improve the performance of their system on \onto. Our results confirm this, since we use 120 types. We further detail the gains we observed for each sub-collection of texts in the test set. Table~\ref{tab:onto.genre} reveals that major improvements over the model of \cite{chiu2015named} are on noisier collections such as telephone conversations (+3 points) and blogs or newsgroups (+2 points). Those type of texts are characterized by a large set of infrequent words, for which classical embeddings are typically poorly trained. Our approach does not seem to suffer from this problem as severely, as discussed in Section~\ref{sec:motivation}. 
	
	\subsection{Ablation Results}
	In this experiment, we directly compare the \lr{} representation with the \sskip{} word-embedding feature set. In order to maintain a high level of performance, both character and capitalization features are used in all configurations. We want to point out that \lr{} vectors are not adapted during training, contrarily to the \sskip{} embeddings. Similarly to Section~\ref{sec:dev}, we report in Table~\ref{tab:ablation}, for each feature configuration, the average F1 score as well as the standard deviation over five runs.
	
	\begin{table}[h]
		\setlength{\tabcolsep}{2mm} 
		\begin{center} 
			\begin{tabular}{|l|l|l|}
				\hline 
				\textbf{Model} & \textbf{\conll}  & \textbf{\onto}\\ 
				\hline
				\sskip              & 90.52 ($\pm$ 0.18) & 86.57 ($\pm$ 0.10) \\
				\lr & 89.94 ($\pm$ 0.16) & 85.92 ($\pm$ 0.12) \\
				\hline
				all                     & \bf 91.73  ($\pm$ 0.10) & \bf 87.95 ($\pm$ 0.13) \\
				\hline
			\end{tabular}
			
		\end{center}
		\caption{F1 scores of differently trained systems  on \conll{ } and \onto{ 5.0} datasets. Capitalization (Section~\ref{sec:Caps})  and character features  (Section~\ref{sec:Chars})  are used by default by all models.}
		
		\label{tab:ablation} 
	\end{table}

	We observe that on both \conll{} and \onto{}, the \sskip{} model outperforms our feature vector approach by 0.65 F1 points on average. The difference is not has high as we first expected, especially since the \sskip{} model is adjusted during training, while our representation is not.	Still, \lr{} vectors seem to  encode a large portion of the information needed to model the NER task. Also, it is worth mentioning that our embeddings are trained on 1.3B words compared to 42B for \sskip.
	
	We also observe  that models that use both feature sets significantly outperform other configurations. To confirm that the gains came from our feature vector and not from increasing the number of hidden units, we tested several \sskip{} models by increasing the LSTM hidden layer dimension so that number of parameters is the same as the model with \lr{} vectors. We observed a degradation of performance on both datasets, mostly due to overfitting on the training set. From those results, we conclude that our lexical representation and the \sskip{} one are complementary. 
	
	\section{Related Works}
	\label{sec:related}

	Traditional approaches to NER, like CRF-based \cite{finkel2005incorporating} and Perceptron-based systems \cite{ratinov2009design} have dominated the field for over a decade. They rely heavily on hand-engineered features~\cite{luo-2015} and external resources such as gazetteers. One major drawback of such an  approach is its weak generalization power \cite{lample2016neural}.
	Current state-of-the art systems~\cite{chiu2015named,strubell2017fast} use a combination of Convolutional Neural Networks (CNNs), Bi-LSTMs, along with a CRF decoder. CNNs are used to encode character-level features (prefix and suffix), while LSTM is used to encode word-level features. Finally, a CRF is placed on top of those models in order to decode the best tag sequence. Pre-trained embeddings obtained by unsupervised learning are core features of those models.  In this work, we show that deep NN architectures can also benefit from lexical features, at least when encoded in the compact form we propose. 
	
	\newcite{tran2017named} and \newcite{peters2017semi} propose an alternative approach different from ours. They incorporate LM embeddings that were pre-trained on a large unlabelled corpus as features for NER. These embeddings allow to generate a representation for a word depending on its context. For instance, the LM embeddings of the word \textit{France} in \enquote{\textit{\underline{France} is a developed country}} is different than that in \enquote{\textit{Anatole \underline{France} began his literary career}}. Such embeddings are trained on very large amount of texts. Our feature set is crafted from distant supervision applied to Wikipedia, a much less time-consuming process which we showed to be nevertheless adapted to rare words.		
	\newcite{chiu2015named} used gazetteer features in order to establish state-of-the-art performance on both \conll{ } and \onto{}. They mined DBPedia in order to compile 4 lists of  named-entities that contain over 2.3M entries. We show that \lr{} representations outperform their gazetteer features.

	\section{Conclusion and Future Work}
	\label{sec:conclusion}
	
	We have explored the idea of generating lexical features for NER out of Wikipedia data automatically annotated with fine-grained entity types. We used \wifine{ }\cite{ghaddara2018wifine}, a Wikipedia dump annotated with fine entity type mentions, for training a vector space that jointly embeds words and named-entities. This vector space is used to compute a 120 dimensional vector per word, which encodes the similarity of the word to each of the entity types. Our results show that our proposed lexical representation, even though it is not adjusted at training time, matches state-of-the-art results compared to more complex approaches on the well-studied \conll{} dataset, and delivers a new state-of-the-art F1 score of 87.95 on the more diversified \onto{} dataset. We further observe larger gains  on collections  with more unfrequent words. 

	The source code and the data we used in this work are publicly available at~\url{http://rali.iro.umontreal.ca/rali/en/wikipedia-lex-sim}, with the hope that other researchers will report gains, when using our lexical representation.	As a future work, we want to investigate the usefulness of our \lr{} feature representation on other NER tasks, including NER in tweets where  out-of-vocabulary and low-frequency words represent a challenge; as well as finer-grained NER which suffers  from the lack of manually annotated training data. 
	
	\section*{Acknowledgements}
	This work has been partly funded by the TRIBE Natural Sciences and Engineering Research Council of Canada CREATE program and Nuance Foundation. We gratefully acknowledge the support of NVIDIA Corporation with the donation of the Titan Xp GPU used for this research. We thank the anonymous reviewers for their insightful comments.
	
	\bibliography{acl2018}
	\bibliographystyle{acl}

\end{document}